\pdfoutput=1
\documentclass[11pt]{article}

\usepackage[]{acl}

\usepackage{times}
\usepackage{latexsym}

\usepackage[T1]{fontenc}
\usepackage[utf8]{inputenc}

\usepackage{microtype}

\usepackage{listings}
\usepackage{newfloat}
\usepackage{listings}
\usepackage{titlesec}
\usepackage{verbatim}
\usepackage{url}
\usepackage{graphicx}
\usepackage{wrapfig}
\usepackage{amsmath}
\usepackage{amssymb}
\usepackage{color,xcolor,colortbl}
\usepackage{enumitem}
\usepackage{algorithm}
\usepackage{algorithmic}
\usepackage[font={small}]{caption}
\usepackage{bm,bbm}
\usepackage{booktabs}
\usepackage{mathtools}
\usepackage{array}
\usepackage{multirow}
\usepackage{soul}
\usepackage{subcaption}
\usepackage{pbox}
\usepackage{pifont}
\usepackage{arydshln}
\usepackage{comment}
\usepackage{array}

\newcommand{\MODELNAME}[0]{\textsl{SPECTRUM}}

\title{\MODELNAME: Speaker-Enhanced Pre-Training for Long Dialogue Summarization}

\author{
    Sangwoo Cho, Kaiqiang Song, Chao Zhao$^\dagger$, Xiaoyang Wang, Dong Yu \\
    Tencent AI Lab, Seattle\\
    UNC Chapel Hill$^\dagger$\\
    \{riversong, shawnxywang, swcho, dyu\}@global.tencent.com, zhaochao@cs.unc.edu$^\dagger$
}

\begin{document}
\maketitle
\begin{abstract}
Multi-turn dialogues are characterized by their extended length and the presence of turn-taking conversations. Traditional language models often overlook the distinct features of these dialogues by treating them as regular text. In this paper, we propose a speaker-enhanced pre-training method for long dialogue summarization, which leverages the inherent structure of multiple-turn dialogues. To support our study, we curate a diverse dataset that includes transcripts from real-world scenarios, movie or TV show transcripts, and dialogues generated by a Large Language Model. We then perform a pre-training, which encompasses the detection of speaker changes, and masked utterance generation. Experimental results of fine-tuned models demonstrate that our model achieves state-of-the-art performance on downstream benchmarks with long context, surpassing baseline models and highlighting the effectiveness of our approach. Our findings highlight the importance of curating pre-training datasets that exhibit diversity and variations in length distribution to ensure effective alignment with downstream datasets.

\end{abstract}

\section{Introduction}
\label{Introduction}

\begin{figure}[h]
\centering
\includegraphics[width=\linewidth]{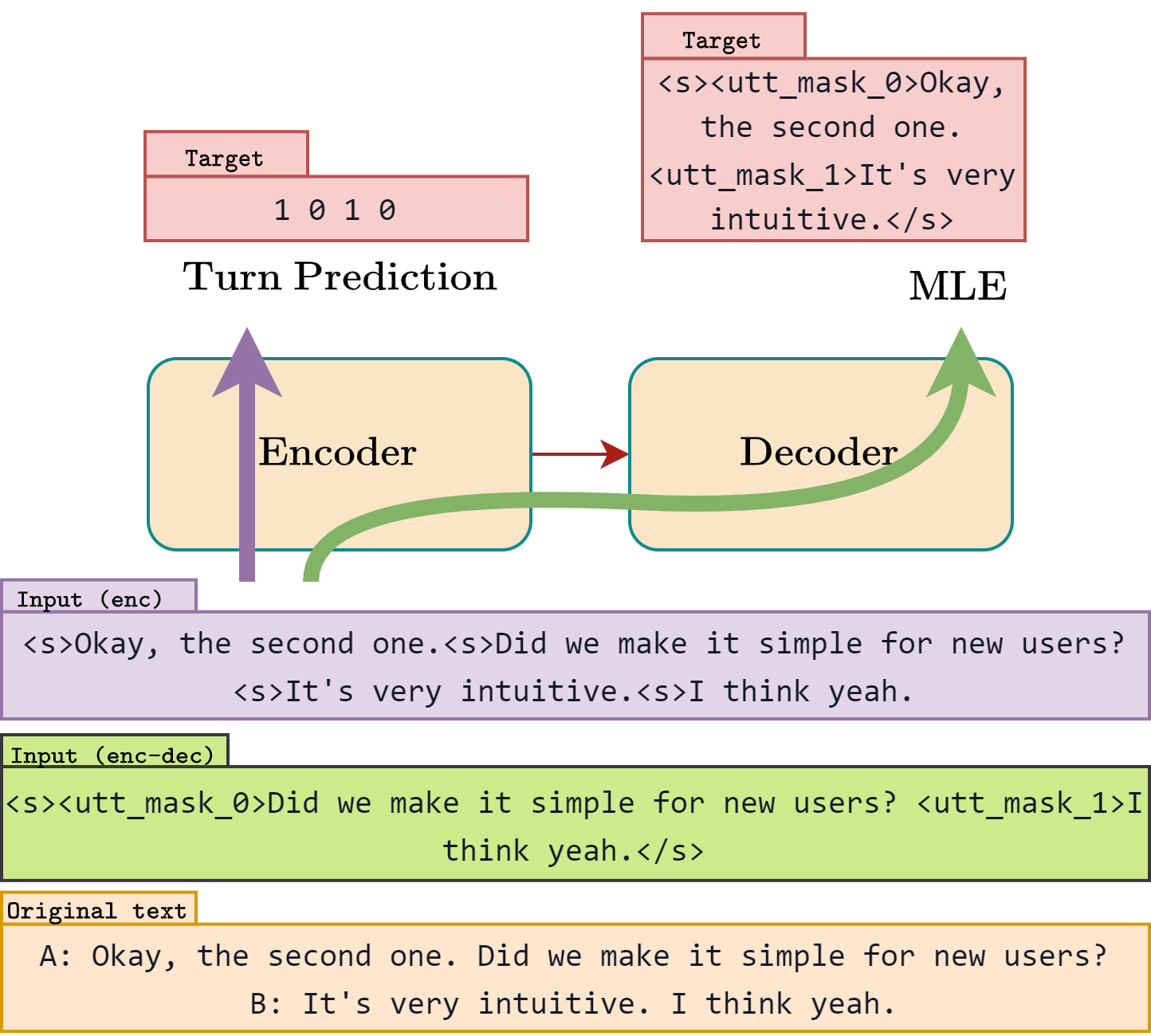}
\caption{The proposed approach implements a speaker-enhanced learning objective within an encoder-decoder model. Two distinct forward paths are incorporated to enhance dialogue understanding. The first path involves only the encoder component, which predicts speaker turn switches. The second path integrates both the encoder and decoder components to generate masked sentences within dialogues, further improving dialogue comprehension.}
\label{fig: model}
\vspace{-0.in}
\end{figure}

Dialogue summarization has been a challenging task in natural language processing.
First, dialogue inherently includes redundant verbal expressions and often excludes certain mentions from the dialogue history.
Consequently, dialogue summarization typically requires a broader context window to effectively handle complex long dependencies.
Second, to gain a thorough comprehension of the content in a conversation, it is essential to take into account the details of the speaker and the signals from the sequential turns of utterances, as these factors significantly influence the interpretation of the information being shared.
Thirdly, dialogue data varies greatly in terms of quality, length, and the presence or absence of speaker information. Finding a method to utilize all these signals is indeed a challenge.

To address the aforementioned issues, we introduce \MODELNAME, %(SPeaker EnhanCed pre-TRaining for long dialogue sUMmarization)
a pre-training approach that incorporates speaker information and leverages the inherent structure of multi-turn dialogues, aiming to enhance the effectiveness of models in interpreting dialogues.
%Unlike the traditional language models which treat the speaker's information as plain text, by incorporating information about the speaker's identity and the turn-based structure of dialogues, our approach enhances the model's ability to capture contextual dependencies and speaker-specific characteristics.
Unlike traditional language models that treat speaker information as plain text, our methodology incorporates the turn-based structure inherent to dialogues.
Integrating speaker information improves the model's ability to focus on speaker-specific traits, allowing for enhanced understanding of long-context dialogues.
%This method enables the model to better understand the nuances and dynamics present in conversations, leading to improved performance in downstream tasks.
% This methodology facilitates the model in comprehending the subtle details and dynamics inherent in dialogues, which consequently leads to an enhancement in performance on subsequent tasks.
To address long-context challenges, we utilize sparse attention networks as our backbone models, specifically LED~\cite{Beltagy2020Longformer} and PEGASUS-X~\cite{Phang2022Pegasusx}.
We collect multi-turn dialogues from diverse sources including two newly collected datasets from YouTube and IMDB Movie Quotes as well as existing public datasets for effective pre-training.
%Additionally, to facilitate the evaluation and analysis of our proposed method, we gathered dialogues from various sources containing multiple turn dialogues including two newly collected datasets from YouTube Channels and Movie Quotes.
%The dataset includes transcripts from real-world scenarios, such as movie or TV show transcripts, as well as dialogues generated by a Large Language Model (LLM).
%This curated dataset provides a comprehensive resource for training and evaluating dialogue understanding models.
We then classify them based on the presence or absence of turn-taking information and perform 2-stage pre-training.
The second stage of pre-training is further tailored based on the diverse length distributions in the datasets.
% The proposed method attains significant performance gains compared to baseline models, underscoring the advantages of integrating speaker-turn information in pre-training and leveraging diverse pre-training data. 

% Through extensive experiments, we demonstrate that our proposed model performs better or on par with state-of-the-art models in the long dialogue summarization task.
% Moreover, we present experimental results that showcase the effectiveness of our speaker-enhanced pre-training method. The proposed method achieves significant performance gains compared to baseline models, highlighting the benefits of incorporating speaker and turn-based information in pre-training dialogue models.

Overall, our work contributes to the advancement of dialogue understanding by introducing a speaker-enhanced pre-training method and offering a curated dataset for pre-training. 
We thoroughly analyze the experimental outcomes and provide insights into best practices for pre-training across different downstream datasets. 
The empirical results validate the efficacy of our approach, showcasing its potential to enhance the performance of dialogue understanding models in various natural language processing applications.
Our contributions are summarized as follows:
\begin{itemize}
\vspace{-0.1in}
\setlength\itemsep{0em}
    \item We propose a two-stage multi-task pre-training approach to leverage the inherent multi-turn structure of dialogues.
    \item We curate pre-training data from diverse sources and explore the effects of varying sequence length distribution in the data.
    \item We achieve state-of-the-art performance compared to models of similar sizes on various datasets and elucidate effective pre-training practices for diverse downstream datasets.
\end{itemize}

%- challenges: dialogue is our main communication tool and is different from the written document as it 
%- long dialogue requires 
%- There are many models 

%Dialogue understanding has been a challenging task in natural language processing (NLP) due to the complex and dynamic nature of conversational interactions. In this paper, we propose a speaker-enhanced pre-training method that leverages the inherent structure of multiple turn dialogues, aiming to improve the performance of dialogue understanding models.

%Our first contribution is the introduction of a speaker-enhanced pre-training method. By incorporating information about the speaker's identity and the turn-based structure of dialogues, our approach enhances the model's ability to capture contextual dependencies and speaker-specific characteristics. This method enables the model to better understand the nuances and dynamics present in conversations, leading to improved performance in downstream tasks.

\section{Related Work}
\label{related_work}

\noindent\textbf{Speaker Information in Dialogue}.
Incorporating speaker information in dialogues has been explored in numerous studies.
Leveraging contextual models to accommodate distinct speaker roles is a practical approach for enhancing the recognition of role patterns and aiding in the comprehension of dialogues~\citep{chi-etal-2017-speaker,Liu2020FillingTG,liu-etal-2020-speaker}. 
Furthermore, ~\citet{Lei2021HierarchicalSS} utilizes the hierarchical model, ~\citet{chi-etal-2017-speaker} is based on the contrastive learning method, and ~\citet{gu2020speakeraware} uses another speaker embedding representation to discern the relationships between speakers and their respective utterances. 
In contrast to solely designing enhanced contextual models, various pre-training methods are employed to improve dialogue comprehension and generation.
For instance, ~\citet{bao-etal-2020-plato} utilizes multi-task pre-training objectives for dialogue response generation and selection, while ~\citet{zhong2022dialoglm} employs diverse masking schemes to effectively capture the dynamics within dialogues.

\noindent\textbf{Efficient Transformers}.
Transformer~\cite{vaswani2017attention} was purposed to solve the sequence-to-sequence tasks.
However, it requires higher computation for longer sequences due to the self-attention operation being quadratic.
Efficient transformers~\cite{yi-2020-efficient} are purposed to reduce the computation complexities while remain the same performance.
In specific, ETC~\cite{ainslie-etal-2020-etc}, BigBird~\cite{zaheer2021bigbird}, and Longformer~\cite{Beltagy2020Longformer} use local attentions with attached global positions to reduce the computation.
The PEGASUS-X~\cite{Phang2022Pegasusx} uses block-wised attention.
The LongT5~\cite{guo-etal-2022-longt5}, on the other hand, implements Transient global attention for the problem.
In this work, we employ Longformer-Encoder-Decoder(LED)~\cite{Beltagy2020Longformer} and PEGASUS-X~\cite{Phang2022Pegasusx} for handling long dialogues.

\noindent\textbf{Pre-training Training Objectives}. 
The training objectives and the corresponding singles are crucial for pre-training the models.
For encoder-only models, BERT~\cite{devlin-etal-2019-bert} first introduces Masked Language Model Loss(MLM).
Subsequently, various models utilize distinct self-supervision signals to enhance the understanding of contextualized semantics in documents. 
For example, ERNIE~\cite{zhang-etal-2019-ernie} incorporates entity knowledge, while SpanBERT~\cite{joshi-etal-2020-spanbert} utilizes span boundaries.
For the decoder-only model, GPT~\cite{Radford2018ImprovingLU,radford2019language} uses language model loss, which predicts the next tokens conditional on all previous tokens.
Similar to the decoder-only model, encoder-decoder models employ conditional language model loss, in which the output token is also conditional on input tokens.
More specifically, BART~\cite{lewis-etal-2020-bart} uses denoising learning objectives to let the model reconstruct the original text from the polluted input text.
PEGASUS~\cite{pegasus-v119-zhang20ae}, introduces sentence masking loss on the decoder side.
T5~\cite{raffel2020exploring-t5} uses a complementary mask for the input and output.
In this paper, we combined turn prediction and masked sentence generation as learning objectives for pre-training.

\noindent\textbf{Long Dialogue Summarization}. 
Typical Summarization dataset like SAMSum~\cite{gliwa-etal-2019-samsum} contains only 94 input words on average.
Similarly, DialogSum~\cite{chen-etal-2021-dialogsum} contains only 137 input words on average.
Unlike typical dialogue summarization datasets, long dialogue summarization data is way harder to collect.
Thus long dialogue summarization data is usually on a small scale.
QMSum~\cite{zhong-etal-2021-qmsum} contains 1,808 summaries for meetings with an average of 9,161 input words.
AMI~\cite{McCowan2005TheAM} contains 137 summaries with around 6 thousand input words.
ICSI~\cite{Janin2003TheIM} contains 59 summaries with more than 13 thousand averaged input words.
SumScreen~\cite{chen-etal-2022-summscreen} contains an average of 5 to 6 thousand input tokens.
In this study, our primary focus is on dialogues or meetings that extend beyond a few thousand words.

%- BookSum~\cite{kryscinski2021booksum}: written document
%- SummScreen~\cite{chen-etal-2022-summscreen}: dialogue
%- QMSum~\cite{zhong-etal-2021-qmsum}, MeetingBank~\cite{hu2023meetingbank}: meeting
%- SCROLLS~\cite{shaham-etal-2022-scrolls} benchmark: natural language tasks with long inputs
%- DialogLM~\cite{zhong2022dialoglm}

\section{Pre-training Objectives}
\label{pretraining_objectives}

We formulate three learning objectives: Speaker Turn Prediction, Masked Language Modeling, and Masked Sentence Generation.
Speaker Turn prediction enhances overall dialogue understanding, while MLM of coreferences, named entities, and verbs aims to enhance entity-level contextual representations in dialogues.
Masked Sentence Generation serves the dual purpose of facilitating dialogue understanding and fostering the ability to generate coherent contextual content.
Our pre-training methodology comprises two pathways: the first exclusively updates the encoder part of the model, whereas the second updates the entire encoder-decoder model.
This pre-training approach improves the model's understanding of contextual dialogues and its ability to generate high-quality sequences aligned with the provided context.

% Our approach incorporates pre-training strategies inspired by PEGASUS summarization pre-training. 
% We further investigate the use of utterance masking and principal token masking techniques specifically tailored for dialogue data. 
% Additionally, we introduce a novel pre-training objective, Speaker Turn Prediction (STP), which aims to enhance dialogue understanding capabilities.

\subsection{Sentence Generation}

We explore two masking strategies in pre-training for models to learn the context of dialogues: word-level and sentence-level.
Initial experiments using the masking methods are conducted and the results are presented in Table~\ref{analysis_pretain_corpus}.
Based on the observation of the superior sentence-level masking strategy and drawing inspiration from the gap sentence generation method~\cite{pegasus-v119-zhang20ae}, we choose to employ sentence-level masking from a single speaker.
% Preliminary experiments using the masking methods are conducted and the results are presented in Table~\ref{analysis_pretain_corpus}.
% We conduct preliminary experiments using the masking methods a random word masking strategy and show the initial results in Sec.~\ref{analysis_pretrain_objective}
% To ensure comprehensive context understanding in dialogues, a simple approach of randomly masking words for sequence-to-sequence self-supervised objectives would only yield limited comprehension of the dialogue's overall context. 
% We conduct preliminary experiments using a random word masking strategy and show the initial results in Sec.~\ref{analysis_pretrain_objective}.
% Taking inspiration from the gap sentence generation method~\cite{pegasus-v119-zhang20ae}, we opt to employ utterance-level masking from a single speaker and concatenate them to form a target output. 
In Figure ~\ref{fig: model}, represented by the green arrow, randomly selected sentences are replaced by $\texttt{<utt\_mask>}$ tokens. 
The target sequence comprises sequences of mask tokens preceding the selected sentences.
% This objective benefits both dialogue understanding and performance in downstream tasks.
Given the dialogues with masked tokens $m$, the objective is to generate the token sequence in the original sentence $t = [w_1, w_2, \dots]$ for $m$, where $w_i$ is the $i$-th token in $t$.
The loss of the generation step is computed as the negative log-likelihood of the conditional language generation:
{\medmuskip=1mu
\thinmuskip=1mu
\thickmuskip=1mu
\nulldelimiterspace=1pt
\scriptspace=1pt
\begin{equation}
\begin{split}
L_{gen} = -\sum_i \log {\rm Pr} (w_i \ | \ w_1, \cdots, w_{i-1}; m)
\end{split}
\label{eq:loss_gen}
\end{equation}}

\subsection{Speaker Turn Prediction}

Dialogues, being a natural form of conversation driven by multiple speakers, present an opportunity to enhance the model's understanding of the contextual nuances inherent in such interactions.
In this study, we aim to teach the model to develop a more comprehensive understanding of dialogue context by explicitly providing speaker-turn information.
By leveraging insights from previous works~\cite{liu-etal-2022-end, cho-etal-2022-toward}, which have demonstrated the benefits of segmenting articles for the summarization task, our model learns to predict turn switches as a means of enhancing dialogue comprehension.
In Figure~\ref{fig: model}, illustrated by the purple arrow, we prepend $\texttt{<s>}$ tokens to each sentence and utilize the resulting contextualized representations for turn switch prediction. 
To facilitate this learning objective, a binary classification approach is employed, leveraging the representations from the top encoder layer. 
{\medmuskip=1mu
\thinmuskip=1mu
\thickmuskip=1mu
\nulldelimiterspace=1pt
\scriptspace=1pt
\begin{equation}
\begin{split}
\hat{y}_{i} = \sigma ( \mathbf{w}^{\top} \mathbf{h}_{i} + b )
% \numberthis\label{eq:pred_seg}
\end{split}
\label{eq:pred_turn}
\end{equation}}
where $\{\mathbf{h}_{i}\}_{i=1}^N$ is contextualized output representations for $N$ sentences in the input dialogues.
$\hat{y}_{i}$ is the turn prediction score and $y_{i}$ is the ground-truth turn switch label, which can be readily obtained from the dialogues.
{\medmuskip=1mu
\thinmuskip=1mu
\thickmuskip=1mu
\nulldelimiterspace=1pt
\scriptspace=1pt
\begin{equation}
\begin{split}
\mathcal{L}_{\textsl{turn}} = - \frac{1}{N} \sum_{i=1}^N \Big( y_{i} \log \hat{y}_{i} + (1-y_{i}) \log (1- \hat{y}_{i}) \Big)\\   
% \quad\quad\quad\quad + (1-y_{i}) \log (1- \hat{y}_{i}) \Big)
\end{split}
\label{eq:loss_turn}
\end{equation}}
This objective can be effectively applied to data containing speaker turns, available in numerous real-world transcripts.

\begin{table*}[!htbp]
\setlength{\tabcolsep}{4pt}
\renewcommand{\arraystretch}{1.1}
\centering
\scriptsize
\begin{small}
\begin{tabular}{|ll|c|c|c|c|c|c|c|c|c|}
\hline
& \textbf{Dataset} & \textbf{Domain} & \textbf{Turn} & \#\textbf{Turn} & \#\textbf{Wds(Total)} & \#\textbf{Wds} & \#\textbf{Insts} & \#\textbf{Wds (M)} & \#\textbf{Insts (M)} \\
\hline
\hline
\multirow{2}{*}{\rotatebox[origin=c]{90}{\textrm{S1}}} & OpenSubtitles  & Movie, TV show & X & - & 2.857B & - & - & 3146 & 908,068 \\
        & Gutenburg-dialogue & Books & X & - & 295M & 130 & 2,279,927 & 3400 & 86,882 \\
\hline
\hline
\multirow{5}{*}{\rotatebox[origin=c]{90}{\textrm{S2}}} & MediaSum  & Interview & O & 28.7 & 689M & 1554 & 443,596 & 1088 & 633,547 \\
        & YouTube-news  & News & O & 16.4 & 50M &  759 & 66,312 & 923 & 54,495 \\
        & YouTube-shows  & Talk shows & O & 42.4 & 26M & 870 & 30,128 & 889  & 29,461 \\
        & Movie Quotes  & Quotes & O & 2.9 & 45M & 38 & 1,181,196 & 918 & 49,398 \\
        % & Summscreen & Screenplay & O & 329 & 128M & 5673 & 22,588 & 1143 &  112,102\\
        & Soda & Social dialogue & O & 7.6 & 146M & 123 & 1,191,132 & 955 & 153,306 \\
\hline
\end{tabular}
\end{small}
\vspace{-0.05in}
\caption{Statistics of pre-training datasets. 
$\texttt{S1}$ and $\texttt{S2}$ represent datasets with and without turns, respectively.
\textbf{\#Wds} is the abbreviation for \textit{number of words}.
\textbf{\#Insts} is the abbreviation for \textit{number of samples}.
$\textbf{\texttt{(M)}}$ indicates the corresponding statistics after packing within each dataset. 
The max number of tokens is 4096 for S1 and 1400 for S2.
}
\label{tab:pretrain_dataset}
\vspace{-0.15in}
\end{table*}

\subsection{Masked Language Modeling (MLM)}

% In line with the BERT~\cite{devlin-etal-2019-bert} methodology, we adopt a token selection approach where 15\% of tokens within the input dialogues are chosen. 
% These selected tokens are then replaced with a $\texttt{[mask]}$ token 80\% of the time or with a random token 10\% of the time. The remaining 10\% of tokens remain unchanged. 

We follow the practice of the BERT~\cite{devlin-etal-2019-bert} methodology.
It is worth noting that dialogues often contain a higher frequency of pronouns, and important information can often be inferred from named entities or verbs.
To utilize this observation, we preprocess the datasets by annotating coreferences, named entities, and verbs as candidate mask tokens\footnote{\texttt{en\_core\_web\_sm} from SpaCy~\citep{spacy2} with \texttt{neuralcoref}~\citep{clark-manning-2016-deep}}.
Initially, we apply MLM with two additional pre-training objectives. 
Nevertheless, following preliminary experiments (Sec.~\ref{analysis_pretrain_objective}), we observe that MLM results in detrimental performance in downstream tasks.
Consequently, we opt to exclude MLM from the final pre-training objective for our model, \MODELNAME.

% Initially, we incorporate  Masked Language Modeling (MLM) alongside two other pre-training objectives. 
% However, based on preliminary experiments (Sec.~\ref{analysis_pretrain_objective}), we found that MLM does not contribute to improved performance in downstream tasks, and decided to exclude MLM from the final pre-training objective for our model, \MODELNAME.

% Our final model is shown in Figure~\ref{fig: model}.
The final pre-training objective is as follows:
{\medmuskip=1mu
\thinmuskip=1mu
\thickmuskip=1mu
\nulldelimiterspace=1pt
\scriptspace=1pt
\begin{equation}
\begin{split}
\mathcal{L}(\Theta) = \mathcal{L}_{\textsl{gen}} + \beta \mathcal{L}_{\textsl{turn}}
\end{split}
\label{eq:loss_all}
\end{equation}}
where $\beta$ is a coefficient that balances sentence-level cross-entropy losses and masked sentence generation and $\Theta$ is the model parameters.

\section{Datasets}
\label{datasets}

\begin{table}[tbp]
\setlength{\tabcolsep}{3pt}
\renewcommand{\arraystretch}{1.1}
\centering
\scriptsize
\begin{small}
\begin{tabular}{|c|c|c|c|c|}
\hline
\textbf{Dataset} & \textbf{Max Target Tks} & \textbf{\#Tks-avg} & \textbf{min} & \textbf{max}  \\
\hline
\hline
S1  &  4096 & 4088.8 & 4080 & 4096 \\
S2-short  & 1400 & 489.7 & 23 & 1400 \\
S2-long  & 1400 & 1388.5 & 1381 & 1400 \\
S2-both  & 1400 & 724.2 & 23 & 1400 \\
\hline
\end{tabular}
\end{small}
\vspace{-0.05in}
\caption{Statistics of various pre-training datasets. 
\textbf{\#Tks} indicates the number of tokens.
}
\label{tab:pretrain_data_stat}
\vspace{-0.15in}
\end{table}

\subsection{Pre-training Corpus}
\label{pretrain_corpus}

For our pre-training process, we utilize two sets of datasets: Stage1 (\textrm{S1}) without turn information; and Stage2 (\textrm{S2}) with turn information, as outlined in Table~\ref{tab:pretrain_dataset}.
The \textrm{S1} data comprises the following two datasets:

\vspace{0.05in}
\noindent\textbf{{OpenSubtitles}}~\cite{lison-tiedemann-2016-opensubtitles2016} This dataset consists of subtitles extracted from movies and TV shows. The English portion contains an extensive corpus of over 2.8 billion words. % \quad

\vspace{0.05in}
\noindent\textbf{{Gutenberg-dialogue}}~\cite{csaky-recski-2021-gutenberg} This dataset consists of dialogues extracted from public-domain books available through Project Gutenberg. %It contains 14.8M utterances and 295M words.

The \textrm{S2} dataset consists of five diverse datasets, including:

\vspace{0.05in}
\noindent\textbf{{MediaSum}}~\cite{zhu-etal-2021-mediasum} This dataset comprises transcripts from media interviews, specifically from CNN and NPR. Speaker information is provided alongside corresponding utterances.

\vspace{0.05in}
\noindent\textbf{{YouTube-news and YouTube-shows}} We have collected the datasets by crawling a variety of YouTube channels, such as news, sports news for YouTube-news; and variety shows, and talk shows for YouTube-shows. Turn switch signals are obtained from the transcripts.

\vspace{0.05in}
\noindent\textbf{{Movie Quotes}} We have collected a dataset of 1.2 million memorable movie quotes from IMDB. This dataset includes the characters, utterances, and stage directions for each quote.

\vspace{0.05in}
\noindent\textbf{{Soda}}~\cite{kim2022soda} This dataset consists of socially-grounded dialogues distilled from GPT-3.5~\cite{ouyang2022training}, a large language model.

To enhance the model's capability to process longer inputs, we pack instances in each dataset up to a maximum of 4096 tokens for \textrm{S1} and 1400 tokens (the average token count across all \textrm{S2} datasets) for \textrm{S2}.
The packed \textrm{S2} dataset is referred to as \textrm{S2-long}.
Nevertheless, we posit that consolidating multiple contexts from different dialogues into a single sequence may introduce confusion to the model.
Therefore, we create a \textrm{S2-short} dataset concatenating all \textrm{S2} datasets without packing to attain coherent context in each sequence.
Finally, we construct \textrm{S2-both} by concatenating \textrm{S2-short} with the long-context \textrm{S2-long} dataset to incorporate diverse input sequence lengths.
The statistics of each pre-training dataset are presented in Table~\ref{tab:pretrain_data_stat}.
Note that we limit the max length of \textrm{S2} datasets constrained to 1400, leading to the truncation of sequences in \textrm{S2-short} and \textrm{S2-both} that surpass this limit.
Since the \textrm{S1} dataset lacks turn information, the speaker turn prediction objective is exclusively applied to the \textrm{S2} dataset.
The experimental results of various pre-training datasets on downstream datasets are presented in Sec.~\ref{analysis_pretrain_objective}.

% We conduct experiments with the various pre-training datasets to analyze their impact on downstream performance (Section~\ref{analysis_pretrain_objective}). Given that the \textrm{S1} data lacks turn information, the speaker turn prediction objective is exclusively applied to the \textrm{S2} data during pre-training.

% We hence consider a \textrm{S2-short} dataset that concatenates all the \textrm{S2} datasets without packing.
% Lastly, the \textrm{S2-both} dataset is considered by concatenating \textrm{S2-short} and \textrm{S2-long} to have more variations in input sequence length.
% These concatenated datasets are referred to as \textrm{S1} and \textrm{S2-long}, respectively.
% The final \textrm{S1} and \textrm{S2} datasets are created by concatenating all the datasets within each category. 
% However, for \textrm{S2}, merging multiple contexts in a single sample can introduce confusion to the model. 
% Hence, we also create an \textrm{S2-short} dataset that concatenates all the \textrm{S2} datasets without merging.
% Consequently, the final pre-training corpus comprises three datasets: \textrm{S1-long} and \textrm{S2-short}, and \textrm{S2-long}. 
% As the \textrm{S1} data lacks turn information, the speaker turn prediction objective is exclusively applied to the \textrm{S2} data during pre-training.

\subsection{Summarization Datasets}
\label{summarization_datasets}

Table~\ref{tab:summ_dataset} shows the statistics for the downstream datasets.
% We choose them since they are dialogues and have long input that can be a good downstream dataset to verify pre-training strategy.
% The choice of the dialogue datasets is motivated by their long input. 
% We choose long dialogue datasets since they are inherently complex and require long context understanding, which makes them suitable for evaluating the effectiveness of our pre-training strategy.
We select long dialogue datasets in different domains given their inherent complexity and need for broad contextual understanding, suitably evaluating our pre-training strategy's effectiveness.

\vspace{0.05in}
\noindent\textbf{{AMI and ICSI}}~\cite{McCowan2005TheAM, Janin2003TheIM} The meeting scripts used in datasets were generated by an Automatic Speech Recognition (ASR) system. The AMI corpus was collected from product design meetings in a company, while the ICSI corpus was collected from academic group meetings in a school. 
% The word error rate (WER) for the AMI corpus was 36\%, and the WER for the ICSI corpus was 37\%.

\vspace{0.05in}
\noindent\textbf{{QMSum}}~\cite{zhong-etal-2021-qmsum} is a dataset specifically designed for query-based meeting summarization. It encompasses meetings from three domains: AMI, ICSI, and committee meetings of the Welsh Parliament and Parliament of Canada. Each query and sample is written by domain experts.

\vspace{0.05in}
\noindent\textbf{{SummScreen}}~\cite{chen-etal-2022-summscreen} consists of community-contributed transcripts of television show episodes obtained from The TVMegaSite, Inc. (TMS) and ForeverDream (FD). Each transcript is accompanied by a summary, which is either a recap sourced from TMS or a recap of the FD shows sourced from Wikipedia and TVMaze.

\begin{table}[tbp]
\setlength{\tabcolsep}{3.2pt}
\renewcommand{\arraystretch}{1.1}
\centering
\scriptsize
\begin{small}
\begin{tabular}{|c|c|c|c|c|}
\hline
\textbf{Dataset} & \textbf{Domain} & \#\textbf{Dialogues} & \#\textbf{Wds} & \#\textbf{Wds (S)} \\
\hline
\hline
QMSum  & Meetings & 1,808 & 9161.1 & 69.7 \\
AMI  & Meetings & 137 & 6007.7 & 296.6 \\
ICSI & Meetings & 59 & 13317.3 & 488.5 \\
SS/FD  & TV show & 4,348 & 5930.2 & 101.2\\
SS/TMS  & TV show & 22,503 & 4599.6 & 344.2 \\
\hline
\end{tabular}
\end{small}
\vspace{-0.05in}
\caption{Statistics of summarization datasets based on dialogues. 
\textbf{\#Wds} is the abbreviation for \textit{number of words}.
$\textbf{\texttt{(S)}}$ indicates words in target summary.
The input and target length are averaged across each dataset.
}
\label{tab:summ_dataset}
\vspace{-0.15in}
\end{table}

\section{Experimental Setup}
\label{experimental_setup}

In our study, we employ sparse Transformer models, specifically LED~\cite{Beltagy2020Longformer} and PEGASUS-X~\cite{Phang2022Pegasusx}, to handle long input dialogues efficiently. % and take advantage of pre-trained models.
Considering the considerable computational requirements for pre-training, we employ the pre-trained checkpoint from PEGASUS-X (568M) and LED-large model (459M).
The experiments are performed using the PEGASUS-X model unless stated otherwise.

\subsection{Pre-training}

In our pre-training process, we follow the general recipe outlined in PEGASUS~\cite{pegasus-v119-zhang20ae} for sentence generation. 
Our pre-training involves two stages, each utilizing distinct datasets.
In the first stage, we utilize the \textrm{S1} data, which contains approximately 3.3 times more words than the \textrm{S2} data. 
This facilitates effective adaptation of the pre-trained models to dialogue data.
The maximum lengths for input and output sequences are set at 4096 and 512 tokens, respectively, with an utterance masking ratio of 12\% for effective training.
We allocate 1\% of the data for the validation set to monitor the training progress.

The second stage involves dual-path training, one for the encoder-only and the other for the encoder-decoder, to compute the joint loss (Eq. \ref{eq:loss_all}) on the \textrm{S2} datasets.
This approach allows for effective learning by improving the model's ability to understand dialogue sequences.
% For the \textrm{S2-short} data, which encompasses a wide range of lengths from different datasets, we truncate tokens longer than 1400 to align with the \textrm{S2-long} data. 
In this stage, we employ a maximum of 1400 input tokens and 256 output tokens, maintaining a masking ratio of 18\% to ensure that the ratio remains below 20\%.
% By following this two-stage pre-training process with specific configurations for each stage, we aim to optimize the model's ability to handle long dialogue contexts, improve dialogue understanding, and align better to the downstream task.
% We use a 1\% subset from the YouTube and Movie Quotes datasets for validation while employing a standard validation set for the remaining datasets.
We employ 1\% data from the YouTube and Movie Quotes datasets for the validation set, while employing a standard validation set for the remaining datasets.

\subsection{Fine-tuning}

We evaluate our pre-trained models by fine-tuning on the downstream summarization datasets as listed in Table~\ref{tab:summ_dataset}.
Considering the average word counts for each dataset, we fine-tune \MODELNAME with an input length of up to 16,384 tokens for QMSum and ICSI, and 8,192 tokens for the remaining datasets.
Due to the limited maximum input length (4096 for \textrm{S1} and 1400 for \textrm{S2})  used in pre-trining, the parameters of positional embedding are repeated up to the respective maximum input lengths for fine-tuning.
We configure the output length to 256 tokens for QMSum and SummScreen/FD, and 512 tokens for the other datasets.
The turn prediction loss is not employed during the fine-tuning process.
During inference, we employ a greedy search method to produce all summaries, thereby eliminating the necessity to introduce new variables during the decoding process.

We report the Rouge evaluation metric~\cite{lin-2004-rouge} by using the \texttt{rouge-score} library\footnote{\url{https://pypi.org/project/rouge-score/}}.
Our evaluation results include Rouge-1 (R-1), Rouge-2 (R-2), Rouge-L (R-L)\footnote{Rouge-L is computed without considering newlines. The summary-level Rouge-L score, which considers sentence splitting, is reported in other systems.}, and the geometric average of the three Rouge scores (R-G). 

% A comprehensive overview of the hyperparameters used in the fine-tuning process can be found in the Appendix section.

\begin{table*}[ht]
\setlength{\tabcolsep}{3pt}
\renewcommand{\arraystretch}{1.15}
\centering
\begin{footnotesize}
\begin{tabular}{|l|cccc|cccc|cccc|}
\hline
& \multicolumn{4}{c|}{\textbf{\textsl{AMI}}} & \multicolumn{4}{c|}{\textbf{\textsl{ICSI}}} & \multicolumn{4}{c|}{\textbf{\textsl{QMSum}}}\\
\textbf{Model} & \textbf{R-1} & \textbf{R-2} & \textbf{R-L} & \textbf{R-G} & \textbf{R-1} & \textbf{R-2} & \textbf{R-L} & \textbf{R-G} & \textbf{R-1} & \textbf{R-2} & \textbf{R-L} & \textbf{R-G}\\
\hline
\hline
% \rowcolor{gray!10}
% \multicolumn{7}{|c|}{\textbf{\textsl{Abstractive Systems}}} \\
% Longfomer & 54.20 & 20.72 & 51.36* & -    & 43.03 & 12.14 & 40.26* & -         & 31.60 & 7.80 & 20.50* & - \\
PGNet & 42.60 & 14.01 & 22.62 & 23.81    & 35.89 & 6.92  & 15.67 & 15.73     & 28.74 & 5.98 & 25.13* & - \\
BART+SCL & 51.40 & 17.81 & 25.30 & 28.50 & - & - & - & - & - & - & - & - \\
HMNet & 52.36 & 18.63 & 24.00 & 28.61    & 45.97 & 10.14 & 18.54 & 20.52     & 32.29 & 8.67 & 28.17*  & -\\
HAT-BART &  52.27 & 20.15 & 50.57* & -     & 43.98 & 10.83 & 41.36* & -        &-&-&-&- \\
DDAMS & 53.15 & \textbf{22.32} & 25.67 & 31.23    & 40.41 & 11.02 & 19.18 & 20.44     &-&-&-&- \\
Summ$^{N}$ & 53.44 & 20.30 & 51.39* & -     & 48.87 & 12.17 & 46.38* & -         & 34.03 & 9.28 & 29.48* & - \\
DialogLED & 54.80 & 20.37 & 52.26* & -     & 50.11 & 13.23 & 47.25* & -         & 34.50 & 9.92 & 30.27* & - \\
DialogLED$^{\dag}$ & 54.45 & 20.20 & 26.03 & 30.58     & \textbf{50.28} & 13.27 & \textbf{20.16} & 23.78    & 34.09 & 9.91 & 20.34 & 19.01 \\
PGS-X (568M) & (50.57) & (19.53) & (25.14) & (29.17)    & (48.65) & (13.40) & (18.26) & (22.83)   & 33.2 & 9.6 & 21.6 & 19.02 \\
LT5$_{\textsf{LG}}$ (770M) & - & - & - & -    & - & - & - & -   & 35.1 & 12 & 23.3 & 21.41\footnotemark{}  \\
LT5$_{\textsf{XL}}$ (3B)   & - & - & - & -    & - & - & - & -   & 34.9 & 11.8 & 23.5 & 21.31\footnotemark[\value{footnote}] \\
\hline
\hline
\MODELNAME & \textbf{55.12} & {21.72} & \textbf{26.73} & \textbf{31.75}    & 50.04 & \textbf{14.5} & 19.81 & \textbf{24.31}   & \textbf{35.86} & \textbf{12.34} & \textbf{23.53} & \textbf{21.84} \\
\hline
\end{tabular}
\end{footnotesize}
\vspace{-0.05in}
\caption{ROUGE scores on meeting summarization datasets: AMI, ICSI, and QMSum. 
\textbf{R-G} is the geometric average of R-1, R-2 and R-L.
* denote the ROUGE-L scores with sentence split. 
$^{\dag}$ indicates Rouge scores computed using outputs from the author's repository.
The numbers in parentheses represent the fine-tuned results using PEGASUS-X.
Our results are significantly better than the baseline models ($\rho$ < 0.05).
\MODELNAME is the same size of PGS-X.
}
\label{tab:results_ami_icsi_qmsum}
\vspace{-0.15in}
\end{table*}

\footnotetext{https://www.scrolls-benchmark.com/leaderboard}
\begin{table}[htbp]
\setlength{\tabcolsep}{3pt}
\renewcommand{\arraystretch}{1.}
\centering
\begin{scriptsize}
\begin{tabular}{|l|cccc|cccc|}
\hline
& \multicolumn{4}{c|}{\textbf{\textsl{SummScreen/FD}}} & \multicolumn{4}{c|}{\textbf{\textsl{SummScreen/TMS}}} \\
\textbf{Model} & \textbf{R-1} & \textbf{R-2} & \textbf{R-L} & \textbf{R-G} & \textbf{R-1} & \textbf{R-2} & \textbf{R-L} & \textbf{R-G} \\
\hline
\hline
BART-large &  33.82 & 7.48 & 29.07* & -    & 43.54 & 10.31 & 41.35* &  - \\
UniLM    & 33.29 & 6.74 & 28.21* & -      & 44.07 & 9.96 & 41.73* & - \\
Summ$^{N}$ & 32.48 & 6.12 & 27.14* & -      & 44.64 & 11.87 & 42.53* & - \\
DialogLED & 36.70 & 8.68 & 31.38* & -     & \textbf{45.22} & 11.69 & 42.86* & - \\
DialogLED$^{\dag}$ & 36.45 & 8.57 & 19.09 & 18.13     & 44.29 & 10.86 & 18.74 & 20.81 \\
PGS-X & 35.7 & 9.1 & 20.6 & 18.84      & - & - & - & -\\
LT5$_{\textsf{LG}}$ & 35.6 & 9.2 & 21.2 & 19.07      & - & - & - & -\\
LT5$_{\textsf{XL}}$ & \textbf{35.8} & \textbf{9.6} & 21.1 & \textbf{19.35}      & - & - & - & -\\
\hline
\hline
\MODELNAME & 35.77 & {9.54} & \textbf{21.22} & \textbf{19.35}   & {45.01} & \textbf{12.21} & \textbf{19.11} & \textbf{21.90} \\
\hline
\end{tabular}
\end{scriptsize}
% \vspace{-0.05in}
\caption{ROUGE scores on SummScreen: FD (ForeverDreaming) and TMS (TV MegaSite). 
}
\label{tab:results_fd_tms}
\vspace{-0.15in}
\end{table}

% Longfomer & 25.90 & 4.20 & 23.80* & -     & 42.90 & 11.90 & 41.60* & - \\

\subsection{Implementation Details}
% pytorch-lightning? yes
We implement our code with Huggingface-transformers~\footnote{\url{https://huggingface.co/docs/transformers/index}} and Pytorch-lightning~\footnote{\url{https://www.pytorchlightning.ai/index.html}}.

During pre-training, we train PEGASUS-X for 20 epochs with stage 1 and 2 strategies: \textrm{S1} with sentence masking objective, \textrm{S2} with joint objectives.
We utilize 64 V100 GPUs with a total batch size of 256.
We use Adam~\cite{kingma2014adam} as our optimizer with a linearly decreased learning rate schedule with 5\% of warmup.
The maximum learning rate is \textrm{$2e^{-4}$}.
We pre-train models with fp32 for stable training and use gradient\-checkpointing for only stage 1 to save GPU memory.
For the second stage of pre-training, we fix $\beta$=3. We experiment with an adaptive method by updating the parameter $\beta$ at the end of each epoch, based on the average loss ratio between $L_{gen}$ and $L_{turn}$ of all steps. However, this approach underperforms in our experiments.
% learning rate schedule: Pretrain - Adam with linear_schedule_with_warmup 5% / Fine-tune - Adafactor with cosine_schedule_with_warmup
% warmup rates
% optimizer (Adam, betas?)
% gradient clipping

During fine-tuning, we keep the same settings of hyper-parameters except for the optimizer.
For saving GPU memory, we use Adafactor~\cite{shazeer2018adafactor} and apply a cosine learning rate decay schedule with 5\% of warmup steps.

\section{Results and Analysis}
\label{result_analysis}

% In this section, we will first introduce the baseline method in Section~\ref{subsec: baseline}.
% Then, we summarize the results of the downstream summarization task in Section~\ref{subsec: results}.
% Finally, detailed analyses of different model components are presented in Section~\ref{subsec: analysis}.

\subsection{Baselines}
\label{subsec: baseline}

We compare \MODELNAME~with various baselines.

\textbf{PGNet}~\cite{see-etal-2017-pgnet} uses copy mechanism to balance between accurate and novel words for the summarization task. 
\textbf{BART+SCL}~\cite{geng-etal-2022-improving-abstractive} utilizes token, turn, and global-level of supervised contrastive learning on short and long dialogue benchmarks.
\textbf{HMNet}~\cite{zhu-etal-2020-hierarchical} employes hierarchical attention structure. It utilizes cross-domain pre-training for meeting summarization.
\textbf{HAT-BART}~\cite{Rohde2021HierarchicalLF} introduces a new hierarchical attention Transformer-based architecture that outperforms standard Transformers.
\textbf{DDAMS}~\cite{Feng2020DialogueDG-ddams} uses a relation graph to model the utterance interactions in a meeting by using different discourse relation models.
\textbf{Summ$^N$}~\cite{zhang-etal-2022-summn} proposes a simple multi-stage summarization framework by generating a coarse summary in multiple stages and finally producing a fine-grained summary.
\textbf{UniLM}~\cite{dong2019unified} is a pre-trained model that employs a shared Transformer and utilizes self-attention masks to control the prediction context.
\textbf{DialogLED}~\cite{zhong2022dialoglm} proposes pre-training strategies for dialogues based on sparse attention and LED models.
\textbf{PGS-X}~\cite{Phang2022Pegasusx} utilizes block-based local attentions for computational efficiency based on PEGASUS~\cite{pegasus-v119-zhang20ae} to incorporate long input.
\textbf{LT5}~\cite{guo-etal-2022-longt5} integrates the extended efficient attention architecture from ETC~\cite{ainslie-etal-2020-etc} and pre-training strategies from PEGASUS~\cite{pegasus-v119-zhang20ae}.

\begin{table}[tbp]
\setlength{\tabcolsep}{3pt}
\renewcommand{\arraystretch}{1.}
\centering
\begin{footnotesize}
\begin{tabular}{|l|cccc|}
\hline
& \multicolumn{4}{c|}{\textbf{\textsl{QMSum}}}  \\
\textbf{Data \& Mask \& Loss} & \textbf{R-1} & \textbf{R-2} & \textbf{R-L} & \textbf{R-G}  \\
\hline
\hline
% LED-large & 48.65 & 13.40 & 18.26 & 22.83  \\
S1-word masking \scriptsize{($L_{gen}$)} & 27.49 & 6.73 & 18.96 & 15.19 \\
S1-sentence masking \scriptsize{($L_{gen}$)} & 29.94 & 7.42 & 19.42	& 16.28 \\
S2-long \scriptsize{($L_{gen}$)} & 29.50 & 7.45 & 19.19 &  16.16      \\
S2-long \scriptsize{($L_{gen}$+$L_{turn}$)} & 30.24 & 7.47 & 19.39  & 16.36     \\
S2-long \scriptsize{($L_{gen}$+$L_{turn}$+$L_{MLM}$)} & 26.96 & 6.52 & 18.39  & 14.79      \\
\hline
\end{tabular}
\end{footnotesize}
% \vspace{-0.05in}
\caption{
ROUGE scores on QMSum with LED-large, comparing word and sentence masking strategies and combinations of learning objectives ($L_{gen}$, $L_{turn}$, $L_{MLM}$).
}
\label{tab:ablation}
\vspace{-0.15in}
\end{table}

\subsection{Results on Summarization Tasks}
\label{subsec: results}
\noindent\textbf{Meeting Summairzation} Table ~\ref{tab:results_ami_icsi_qmsum} presents a performance comparison of \MODELNAME, which is based on PGS-X, on three long meeting summarization tasks: AMI, ICSI, and QMSum. 
The performance of \MODELNAME~is reported from the best results based on different pre-trained checkpoints. A more detailed analysis of different pre-trained checkpoints is provided in Section ~\ref{analysis_pretrain_objective}.
In all datasets, we observe a significant improvement in Rouge scores with \MODELNAME~compared to PGS-X. The R-G gains for AMI, ICSI, and QMSum are 2.58, 1.48, and 2.82, respectively. 
This highlights the effectiveness of our pre-training strategy and the carefully curated data that aligns with the downstream tasks, resulting in improved performance.
Furthermore, our \MODELNAME~model outperforms other pre-trained models such as HMNet and DialogLED by a large margin. Notably, \MODELNAME~even surpasses larger-sized models with different efficient Transformers, LongT5-large and LongT5-xl, on the QMSum task. This highlights the superior performance and efficiency of our approach in long meeting summarization tasks.

\vspace{0.05in}
\noindent\textbf{Screenplay Summairzation} In Table ~\ref{tab:results_fd_tms}, we present the performance of the \MODELNAME~models on the SummScreen dataset. Our observations indicate that \MODELNAME~surpasses all other models in R-G on both subsets of the dataset. 
Additionally, \MODELNAME~demonstrates competitive performance when compared to the 3B-size model, LongT5-xl. 
These results highlight again the effectiveness of \MODELNAME~in achieving strong performance on the long dialogue summarization dataset.

\subsection{Analysis}
\label{subsec: analysis}

\subsubsection{Effect of Pre-training Objectives}
\label{analysis_pretrain_objective}

A series of ablation studies are conducted to analyze the masking strategies of the generation learning objective and identify the optimal combination of three pre-training objectives.
The effectiveness of various pre-trained models using LED-large is demonstrated through fine-tuning on QMSum, as illustrated in Table~\ref{tab:ablation}.
The first two rows indicate that sentence-level masking outperforms word-level masking with the $L_{gen}$ loss pre-trained on \textrm{S1}.
The Subsequent rows show the result of the complementarity of each learning objective.
We employ the sentence-masking strategy for the $L_{gen}$ loss and set $\beta$ to 1 in the experimental setup.
Notably, the joint computation of the turn prediction learning objective ($L_{turn}$) leads to enhanced performance.
We additionally investigate the impact of the MLM loss ($L_{MLM}$) on downstream performance. 
The last row illustrates a notable degradation in performance. 
We hypothesize that masking tokens in the encoder input introduce confusion, impeding the model's accurate prediction of turns and exacerbating the learning process within the encoder path. 
Consequently, $L_{MLM}$ is excluded from our final model.

\begin{table}[tbp]
\setlength{\tabcolsep}{3pt}
\renewcommand{\arraystretch}{1.1}
\centering
\begin{small}
\begin{tabular}{|l|cccc|}
\hline
& \multicolumn{4}{c|}{\textbf{\textsl{QMSum}}}  \\
\textbf{CKPT} & \textbf{R-1} & \textbf{R-2} & \textbf{R-L} & \textbf{R-G}  \\
\hline
\hline
PGS-X & 33.2 & 9.6 & 21.6 & 19.02  \\
S1   & 35.54 & 11.74 & 23.05 & 21.26 (\textcolor{red}{$\uparrow$}2.24)     \\
S2-short & 34.84 & 10.81 & 21.54 & 20.09 (\textcolor{red}{$\uparrow$}1.07)     \\
S2-short-ckpt & 35.23 & 11.84 & 23.29 & 21.34 (\textcolor{red}{$\uparrow$}2.32)     \\
S2-long & 35.08 & 11.73 & 22.72 & 21.07  (\textcolor{red}{$\uparrow$}2.05)    \\
S2-long-ckpt & 36.06 & 11.52 & 22.95 & 21.20 (\textcolor{red}{$\uparrow$}2.18)     \\
S2-both-ckpt & 35.86 & 12.34 & 23.53 & {21.84} (\textcolor{red}{$\uparrow$}\textbf{2.82})   \\
\hline
\end{tabular}
\end{small}
% \vspace{-0.05in}
\caption{ROUGE scores on QMSum using different pre-trained checkpoints.
}
\label{tab:results_qmsum_ckpt}
\vspace{-0.05in}
\end{table}

% \begin{table}[t]
% \setlength{\tabcolsep}{3pt}
% \renewcommand{\arraystretch}{1.}
% \centering
% \begin{footnotesize}
% \begin{tabular}{|l|cccc|cccc|}
% \hline
% & \multicolumn{4}{c|}{\textbf{\textsl{QMSum}}} & \multicolumn{4}{c|}{\textbf{\textsl{SummScreen/TMS}}} \\
% \textbf{CKPT} & \textbf{R-1} & \textbf{R-2} & \textbf{R-L} & \textbf{R-G} & \textbf{R-1} & \textbf{R-2} & \textbf{R-L} & \textbf{R-G} \\
% \hline
% \hline
% S1   & 35.54 & 11.74 & 23.05 & 21.26      & 44.07 & 9.96 & 41.73* & - \\
% S2-short & 34.84 & 10.81 & 21.54 & 20.09     & 42.90 & 11.90 & 41.60* & - \\
% S2-short-ckpt & 35.23 & 11.84 & 23.29 & 21.34      & 44.64 & 11.87 & 42.53* & - \\
% S2-long & 35.08 & 11.73 & 22.72 & 21.07     & 45.22 & 11.69 & 42.86* & - \\
% S2-long-ckpt & 36.06 & 11.52 & 22.95 & 21.20     & 44.29 & 10.86 & 18.74 & 20.81 \\
% S2-both-ckpt & 35.86 & 12.34 & 23.53 & \textbf{21.84}     & 44.29 & 10.86 & 18.74 & 20.81 \\
% \hline
% \end{tabular}
% \end{footnotesize}
% % \vspace{-0.05in}
% \caption{ROUGE scores on QMSum and SummScreen using different pre-trained checkpoints.
% }
% \label{tab:results_fd_tms}
% % \vspace{-0.15in}
% \end{table}
\begin{table}[tbp]
\setlength{\tabcolsep}{3pt}
\renewcommand{\arraystretch}{1.1}
\centering
\begin{small}
\begin{tabular}{|l|cccc|}
\hline
& \multicolumn{4}{c|}{\textbf{\textsl{ICSI}}}  \\
\textbf{CKPT} & \textbf{R-1} & \textbf{R-2} & \textbf{R-L} & \textbf{R-G}  \\
\hline
\hline
PGS-X & 48.65 & 13.40 & 18.26 & 22.83  \\
S1   & 49.00 & 13.82 & 18.56 & 23.25 (\textcolor{red}{$\uparrow$}0.42)     \\
S2-short & 49.82 & 14.16 & 19.38 & 23.91 (\textcolor{red}{$\uparrow$}1.08)     \\
S2-short-ckpt & 49.00 & 12.63 & 17.75 & 22.23 (\textcolor{blue}{$\downarrow$}0.60)     \\
S2-long & 47.31 & 13.29 & 18.34 & 22.59  (\textcolor{blue}{$\downarrow$}0.24)    \\
S2-long-ckpt & 50.04 & 14.50 & 19.81 & 24.31 (\textcolor{red}{$\uparrow$}\textbf{1.48})     \\
S2-both-ckpt & 48.01 & 12.52 & 17.89 & 22.07 (\textcolor{blue}{$\downarrow$}{0.76})   \\
\hline
\end{tabular}
\end{small}
% \vspace{-0.05in}
\caption{ROUGE scores on ICSI using different pre-trained checkpoints.
}
\label{tab:results_icsi_ckpt}
\vspace{-0.15in}
\end{table}

\begin{figure*}[htbp]
\centering
\includegraphics[width=1\linewidth]{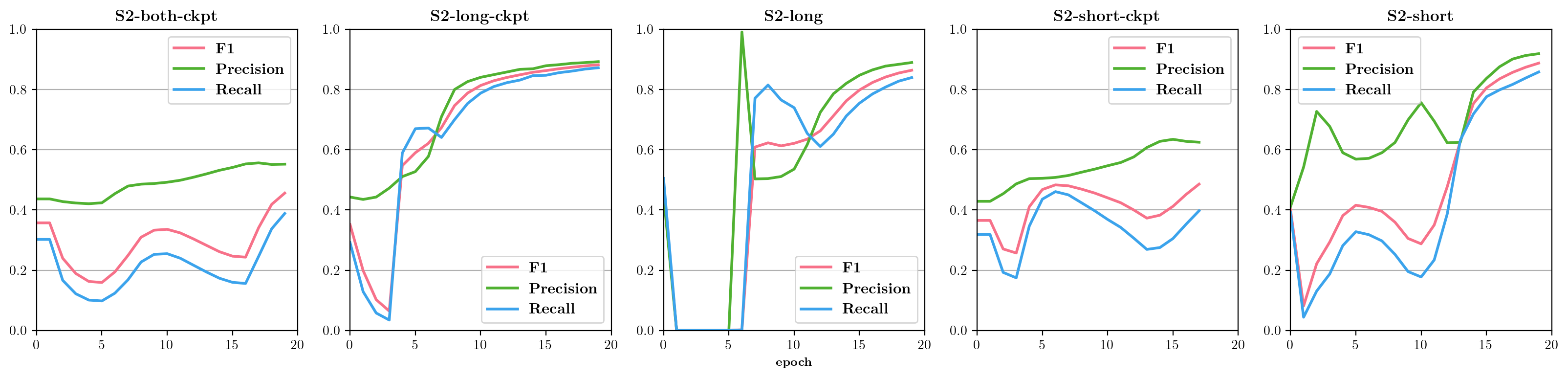}
\caption{F1, precision, recall scores for the turn switch prediction. Each figure shows the performances with the corresponding \textrm{S2} pre-training data with (`-ckpt') or without parameter initialization pre-trained on \textrm{S1}.}
\label{fig:f1pr}
\end{figure*}

\subsubsection{Effect of Pre-training Corpus}
\label{analysis_pretain_corpus}

Table~\ref{tab:results_qmsum_ckpt},~\ref{tab:results_icsi_ckpt} shows the fine-tuned results on QMSum and ICSI datasets using different pre-trained checkpoints. 
PGS-X refers to the original PEGASUS-X checkpoint, \textrm{S1} denotes pre-training on \textrm{S1} with $L_{gen}$, and \textrm{S2} corresponds to pre-training on different \textrm{S2} data with or without initialization from the \textrm{S1} checkpoint.
Significant performance improvement is evident across all pre-trained checkpoints on QMSum. 
However, performance gains are comparatively limited on ICSI. 
We postulate that this phenomenon is attributed to the small size of the ICSI dataset, comprising only 43 training instances, which poses a challenge for the model to align effectively. % with the dialogue understanding capability of the pre-trained model.
In contrast, the QMSum dataset contains 1,257 training samples, allowing the model to better adapt during fine-tuning.
This observation aligns with the insights proposed by ~\cite{pegasus-v119-zhang20ae}, emphasizing the significance of aligning the domain of pre-trained models to facilitate more effective transfer to downstream tasks.

\subsubsection{Effect of Speaker Turn Prediction}

Figure~\ref{fig:f1pr} presents the F1, precision, and recall scores for turn switch prediction concerning various \textrm{S2} data with the initialization of the \textrm{S1} checkpoint. Several noteworthy observations can be derived from the results.
Firstly, it is observed that stage2 pre-training attains high turn prediction performance even without initializing the model with the \textrm{S1} checkpoint, as evidenced by the \textrm{S2-short} and \textrm{S2-long} cases.
% This indicates the effectiveness of stage 2 pre-training in improving turn prediction capabilities.
Furthermore, the \textrm{S2-long-ckpt} configuration exhibits notably high F1 scores. 
We posit that the consistent length of the \textrm{S1} data aligns effectively with the characteristics of the \textrm{S2-long} data, thereby contributing to the observed high prediction performance.

On the other hand, \textrm{S2-short-ckpt} and \textrm{S2-both-ckpt}, which involve different data distributions in terms of length, show degraded prediction performance during pre-training. 
This suggests that the deviation of data length in two-stage training adversely affects the model's ability to accurately predict turns.
Importantly, the model's ability to predict turn switches aligns closely with the downstream results on the ICSI dataset (Table~\ref{tab:results_icsi_ckpt}), which has a limited number of training samples. 
The parameter initialization with a checkpoint trained on a different data length distribution can impair the fine-tuning performance on scarce data, contrasting with the opposite scenario observed with a sufficiently large dataset (Table~\ref{tab:results_qmsum_ckpt}).
% This observation indicates that pre-training has a more significant impact on the performance of finetuning with small-sized datasets.
Taken together, these findings highlight the importance of pre-training methodologies, dataset characteristics, and the alignment between the pre-training and fine-tuning stages. 
% They suggest that optimizing pre-training strategies can significantly enhance performance, especially for downstream tasks involving small training samples.

\section{Conclusion}
\label{conclusion}

In this paper, we have introduced \MODELNAME, a speaker-enhanced pre-training method that specifically addresses the challenges associated with dialogue summarization. 
Our two-stage pre-training approach leverages the natural structure of dialogues with multiple turns, allowing the model to better comprehend the intricacies of dialogue interactions.
To enable robust training of models, we utilized a comprehensive dataset comprising real-world transcripts, including dialogues from movies, TV shows, and dialogues generated by large language models. 
This diverse dataset with turn-taking information provides solid signals for training dialogue understanding models.
Our experimental results demonstrate the effectiveness of our speaker-enhanced pre-training method and show significant performance gains over baseline models. 
% We have achieved significant performance gains over baseline models, highlighting the benefits of incorporating speaker and turn-based information in dialogue modeling. 
By capturing and leveraging speaker and turn-based information, our approach enhances the ability of models to generate accurate and coherent summaries of dialogues.
Moving forward, future research endeavors can further explore and refine our pre-training methods. 
% Additionally, investigating additional techniques for capturing and leveraging speaker and turn-based information in dialogue understanding tasks holds promise for advancing the field. 

\section{Limitation}
\label{sec:limitation}
While our work introduces novel learning objectives and exhibits promising results, it is not without limitations.

Although our model is designed for long dialogue summarization, it still encounters difficulties when dealing with exceptionally long dialogues. With the increase in length, the complexity of the dialogue also grows, and it becomes harder to maintain the semantic coherence and continuity of the summarization output. This is a common problem in summarization tasks, but its impact is amplified in long dialogue summarization due to the increased amount of data the model must process and understand.

The current evaluation metrics we employ have their inherent limitations. While ROUGE scores have been widely used in summarization tasks, they may not fully capture the quality of generated summaries, such as the preservation of speaker-specific nuances or shifts in conversation topics. 
Also, given the lengthy dialogues, conducting human evaluation is both challenging and potentially inaccurate. 
The constraint on context length in Large Language Models often poses challenges for evaluating long dialogue summarization, and LLMs may potentially exhibit hallucination problems.
Therefore, further research on better representative evaluation metrics is needed.

Another aspect of the proposed model is open to many potential risks such as loss of context in conversation, contextual inconsistency, biased summary in dialogue, and ethical concerns in social interactions.
Addressing these risks involves continuous improvement of the summarization models, careful consideration of ethical implications, and user feedback to enhance the system's performance in handling diverse and dynamic dialogues. 
Regular updates and refinements to the underlying models can contribute to mitigating these challenges over time.

Lastly, the training and deployment of our model demand significant computational resources. This may limit the use of our method in real-world applications, particularly in scenarios where computational resources are constrained.

Future work will aim to address these limitations. We plan to refine our approach to better handle lengthy and complex dialogues, improve its ability to generalize across various domains, and explore new ways to quantify the performance of our model more effectively.

% \newpage

% \input{section/7-Limitations.tex}

% \input{section/8-Ethical.tex}

% Entries for the entire Anthology, followed by custom entries
\bibliography{custom}  % anthology
\bibliographystyle{acl_natbib}

\appendix

% \section{Example Appendix}
% \label{sec:appendix}
% This is an appendix.

\end{document}